\documentclass[5p,twocolumn,sort&compress]{elsarticle} 
\journal{Neurocomputing}
\usepackage{amsmath,epsfig}
\usepackage[english]{babel}
\usepackage[latin1]{inputenc}
\usepackage{hyphenat}
\usepackage{amssymb,amsthm,amsfonts}
\theoremstyle{definition}

\theoremstyle{definition}

\usepackage{lscape}
\usepackage{bm}
\usepackage{verbatim}
\usepackage{bbm}
\usepackage[table]{xcolor}
\usepackage{setspace}
\usepackage{courier}
\usepackage[T1]{fontenc}
\usepackage{rotating,psfrag}
\usepackage{color}
\usepackage{picins}

\newif\ifpdf
\ifx\pdfoutput\undefined
   \pdffalse
\else
   \pdfoutput=1
   \pdftrue
\fi
\ifpdf
   \usepackage{graphicx}
   \usepackage{epstopdf}
\else
   \usepackage{graphicx}
\fi

\begin{document}

\title{Reinforcement learning based sensing policy optimization for energy efficient cognitive radio networks}
\begin{frontmatter}

\author[aalto]{Jan Oksanen\corref{cor1}}
\ead{jhoksane@wooster.hut.fi}
\author[aalto,princeton]{Jarmo Lundén\fnref{fnJ}}
\ead{jrlunden@wooster.hut.fi}
\author[aalto]{Visa Koivunen}
\ead{visa@wooster.hut.fi}
\cortext[cor1]{Corresponding author}
\fntext[fnJ]{J. Lundén's work has been supported by the Qatar National Research Fund and the Finnish Cultural Foundation.}
\address[aalto]{Aalto University School of Electrical Engineering, SMARAD CoE, Department of Signal Processing and Acoustics, P.O. Box 13000, FI-00076 Aalto, Finland}
\address[princeton]{Princeton University, Department of Electrical Engineering, Princeton, NJ 08544, USA}

\begin{abstract}
This paper introduces a machine learning based collaborative multi-band spectrum sensing policy for cognitive radios. The proposed sensing policy guides secondary users to focus the search of unused radio spectrum to those frequencies that persistently provide them high data rate. The proposed policy is based on machine learning, which makes it adaptive with the temporally and spatially varying radio spectrum. Furthermore, there is no need for dynamic modeling of the primary activity since it is implicitly learned over time. Energy efficiency is achieved by minimizing the number of assigned sensors per each subband under a constraint on miss detection probability. It is important to control the missed detections because they cause collisions with primary transmissions and lead to retransmissions at both the primary and secondary user.
Simulations show that the proposed machine learning based sensing policy improves the overall throughput of the secondary network and improves the energy efficiency while controlling the miss detection probability. 
\end{abstract}

\end{frontmatter}

\section{Introduction} \label{intro}
\begin{figure}[t]
        \centering      \includegraphics[width=0.8\columnwidth]{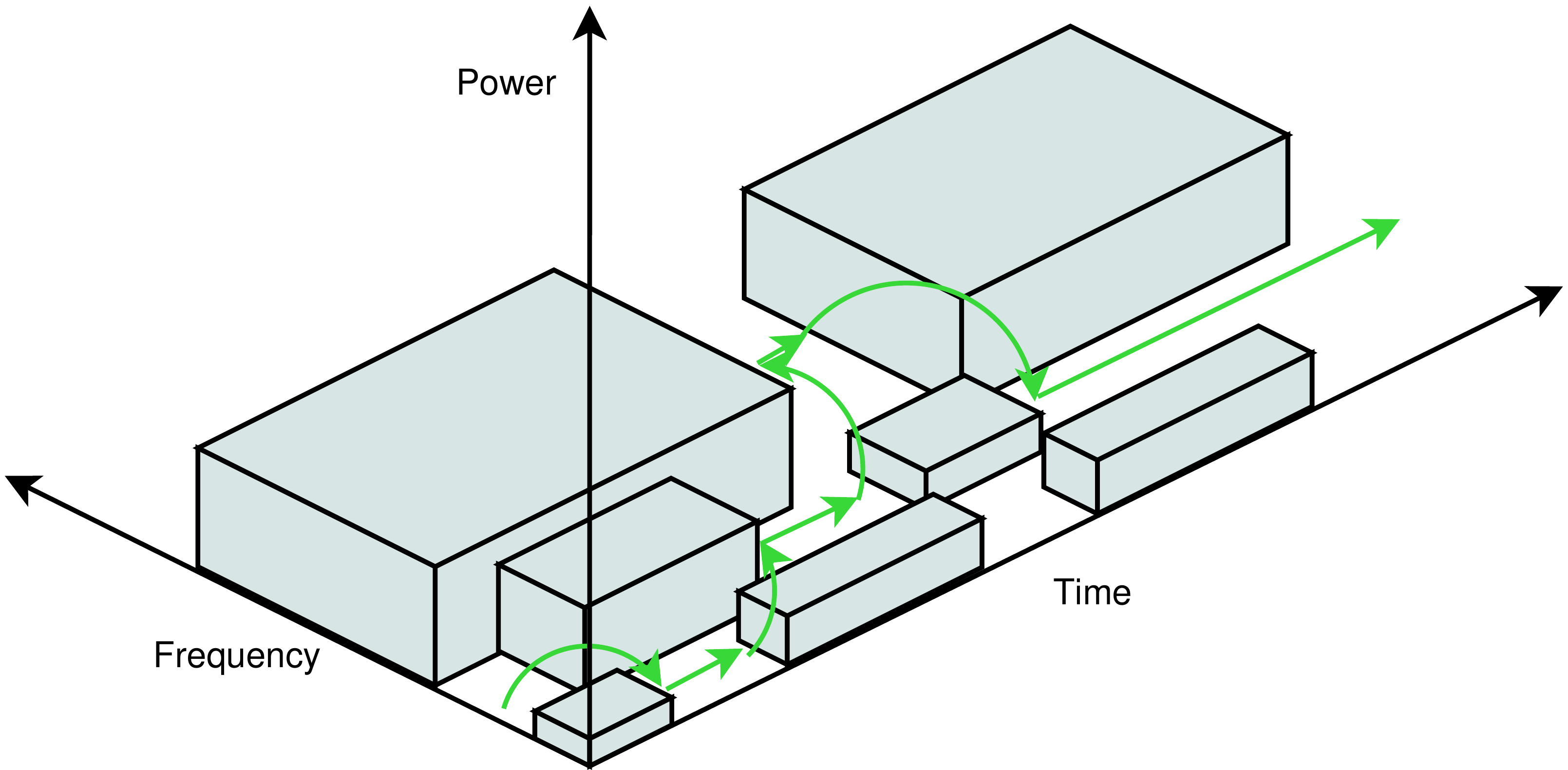}
        \caption{Spectrum holes in time and frequency in a given location. A spectrum hole emerges when a primary user (activity indicated by the blocks) vacates its frequency. Secondary users try to opportunistically detect and access these spectrum holes (indicated by the green arrows).}
        \label{spectrumHoles}
\end{figure}

The increasing demand for wireless services has made the usable radio spectrum a scarce and expensive resource. Part of the scarcity problem are the spectrum allocation policies that do not exploit the fact that the state of the radio frequency spectrum is time and location varying. Measurement campaigns \cite{cabric} have in fact shown that large parts of the spectrum are underutilized because the license holders are not using the spectrum or because the fact that wireless signals attenuate in $2-4$ power of distance is not fully exploited. Underutilized spectrum is time-frequency-location varying resource and radio wave propagation and signal attenuation are important factors in determining where spectrum opportunities or areas of harmful interference occur. Identifying temporal and spatial spectrum holes has been the key motivation behind cognitive radio (CR) and dynamic spectrum access (DSA) \cite{haykin}. Figure \ref{spectrumHoles} illustrates how spectrum holes emerge in time and frequency. 

\begin{figure}[t]
        \centering      \includegraphics[width=0.8\columnwidth]{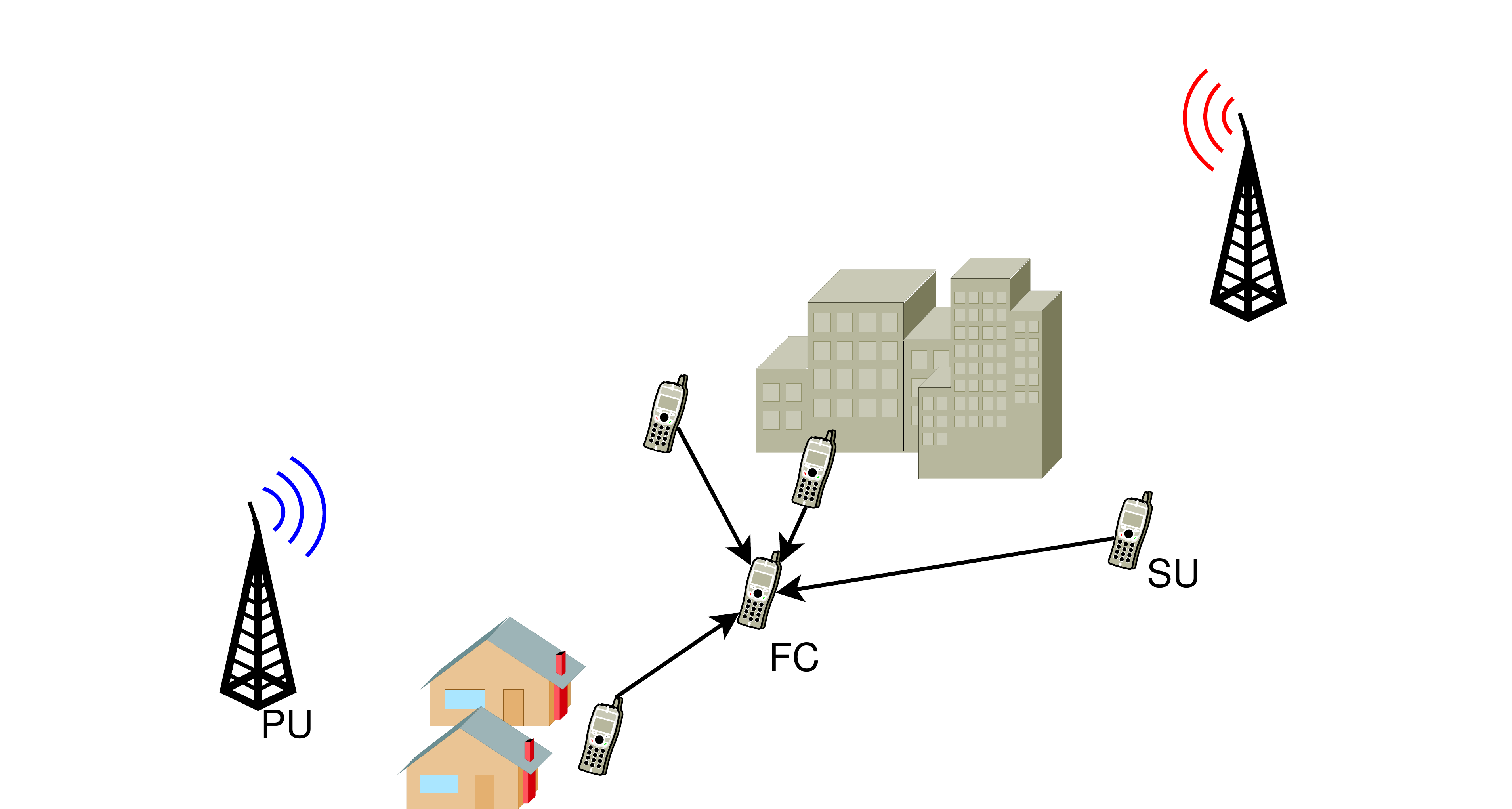}
        \caption{A cognitive radio setting. The SUs are collaboratively sensing whether the PUs are active or not. After sensing the spectrum the SUs send their sensing results to a common fusion center (FC) that makes a global decision about the state of the spectrum and grants access to the spectrum for one of the users if the spectrum is found unoccupied. Cooperative spectrum sensing provides spatial diversity to overcome the effects of slow fading caused by large objects and fast fading caused by multi-path propagation and mobility.}
        \label{CRsettingNCjournal}
\end{figure}

CR systems try to use the licensed radio spectrum in an agile manner while guaranteeing that the licensed users will not be interfered (see figure \ref{spectrumHoles}). A spectrum opportunity is a situation in which secondary users (SU) are able to communicate on a licensed frequency without interfering the primary user (PU) and without being themselves interfered by the PU \cite{zhao}. In order to find such spectrum opportunities CR systems need to sense the spectrum (see figure \ref{CRsettingNCjournal}). 

A CR network can be considered to consist of $N_S$ spatially distributed wireless terminals that identify free frequencies across a wide spectrum of interest that is assumed to have been divided into $N_B$ subbands. In order to mitigate the effects of fading, cooperative detection schemes have been proposed in the literature \cite{haykin,lunden,chaudhari}. This means that a part of the spectrum is simultaneously sensed by multiple SUs that send their local test statistics to a fusion center (FC) which then makes a global decision about the state of the spectrum. With such cooperation, the probability of detection at a given signal-to-noise ratio (SNR) is increased, or for equal performance, simpler detector structures may be employed. 

An important function performed by the FC is the \emph{spectrum sensing policy}, which is also the focus of this paper. A spectrum sensing policy guides the SUs about who is sensing, which part of the spectrum and when. One of the main targets of a sensing policy is to select those frequency bands for sensing that persistently provide more spectrum opportunities and throughput for the SU network.

\subsection{Contribution of the paper}

In this paper a reinforcement learning based multi-user, multi-band spectrum sensing policy is proposed. The proposed sensing policy balances between exploring and exploiting different parts of the radio spectrum and different sensing assignments. It decides which frequency bands to sense as well as which SU is assigned to do the sensing. 
In the exploitation phase the sensing assignment for the high throughput subbands is found by minimizing the number of assigned SUs subject to a constraint on the miss detection probability. Moreover, the probability of false alarm is constrained by using Neyman-Pearson detectors. Minimization of the number of simultaneously sensing SUs improves the energy efficiency of the battery operated SUs. The minimization is formulated as a binary integer programming (BIP) problem that may be solved exactly by a branch-and-bound type algorithm or approximately by using approximative methods such as the iterative Hungarian method considered in this paper. The proposed policy may reduce the number of active sensors up to a factor of $1/D$, where $D$ is the diversity order of a fixed sensing policy. In the exploration phase different pseudorandom sensing assignments with fixed diversity order are explored in order to re-adapt to possible changes in the PU activity and channel conditions. 
On one hand, spatial diversity improves the detector performance in the face of fading and shadowing but on the other hand reduces the number of simultaneously sensed frequency bands by the secondary network. Cognitive network may use multiple idle frequency bands in order to improve rate or reliability of the network.
%

Some preliminary ideas and results related to this paper were presented in \cite{oksanen4}. The contributions of this paper are:
\begin{itemize}
        \item We propose a machine learning based spectrum sensing policy for cognitive radio that:
        \begin{itemize} 
        \item provides high throughput for the SUs,
        \item reduces missed detections,
        \item is energy efficient, 
        \item is adaptive to non-stationary PU behavior and channel conditions.
        \end{itemize}
        \item Analytical expressions for the convergence of the proposed sensing policy in stationary scenarios are derived.
        \item Extensive simulation results highlighting the excellent performance of the proposed sensing policy in various stationary and non-stationary scenarios are shown.
        \item We show that a simple and fast approximative algorithm based on the Hungarian method may be used to find near optimal sensing assignments. 
\end{itemize}
The main difference with this paper and the related work in the literature, in addition to the methodology, is the exploitation of the information about the sensing performances of the SUs to optimize the sensing assignments in an energy efficient manner.  

This paper is organized as follows. In section \ref{relatedWork} the related work to this paper is briefly summarized. The system model of cooperative multi-band sensing is described in section \ref{systemModel}. In section \ref{eGeedy} an energy efficient reinforcement learning based sensing policy is proposed and analytical results on the convergence rate of the $Q$-values in the sensing policy are derived. Section \ref{simulations} shows and discusses the simulation results of the performance of the proposed sensing policy. The paper is concluded in section \ref{conclusions}.

\section{Related work}\label{relatedWork}
The task of choosing which frequency band to sense may be formulated as a restless multi-armed bandit (RMAB) problem. In RMAB problems a player bets on $L$ out of $N$ slot machines ($L\geq 1$, $N\geq L$) targeting to maximize its long term profit. The term restless comes from the fact that also the states of the non-played machines may change; similarly as the state of the not sensed frequency bands may change in a CR setting.
In \cite{zhao,zhao2,zhao3,zhao4,filippi,liu,liu2, liu3} spectrum sensing policies are derived based on the framework of partially observable Markov decision processes (POMDPs). 
In \cite{liu3} a closed form Whittle index policy for perfectly known Markovian reward distributions was derived and shown to be optimal under certain conditions. 

In a case where the player does not have prior knowledge about the reward distributions of the different machines (or as in this case about the throughputs of the different frequency bands), it is obviously impossible to derive optimal action selection policies. In such case machine learning is an attractive approach for solving the problem. A known issue with machine learning methods is the so-called exploitation-exploration trade-off, which emerges when the player has to decide whether to try to exploit the seemingly best machine (or frequency band) at the moment or to explore other machines in hope of finding even better one.
A standard method for tackling multi-armed bandit problems is the Q-learning algorithm \cite{watkins} with $\epsilon$-greedy exploration \cite{sutton98}. 
An alternative way for balancing the trade-off between exploration and exploitation is to use confidence bounds. Namely, in \cite{auer} a simple policy based on upper confidence bounds (UCB) was proposed and shown to reach the optimal regret rate when the rewards are independent and stationary. An UCB policy that suits better for non-stationary rewards was developed in \cite{kocsis}.
In \cite{berthold} a single-user reinforcement learning method was proposed for selecting between 3 future actions: continuing sensing at the current frequency band $b$ and transmitting data, sensing an out-of-band frequency band $\tilde{b}$, and switching the SU system to an out-of-band frequency band $\tilde{b}$. Action selection is done using the softmax method.

\section{System model} \label{systemModel}
The SU network consists of $N_S$ cooperating wireless SU terminals sensing the radio spectrum. The spectrum of interest is assumed to be divided into $N_B$ frequency subbands that may have different bandwidths and may be occupied by different primary operators. The subbands may be scattered in frequency. Depending on the front-end design of the SU device, one SU can sense up to $K_s$ subbands at a time. 

In this paper it is assumed that the SUs cooperate by sending their local binary decisions to a FC, that makes a global decision about the availability of the spectrum for all SUs. This brings spatial diversity and increased scanning speed. Spatial diversity is obtained when multiple SUs sense the same part of the spectrum simultaneously from different locations and then form a global decision. Scanning speed is increased since each SU may get sensing information about up to $\sum_sK_s$ subbands simultaneously. 

\begin{figure}[htp!]
        \centering
                \includegraphics[width=0.9\columnwidth]{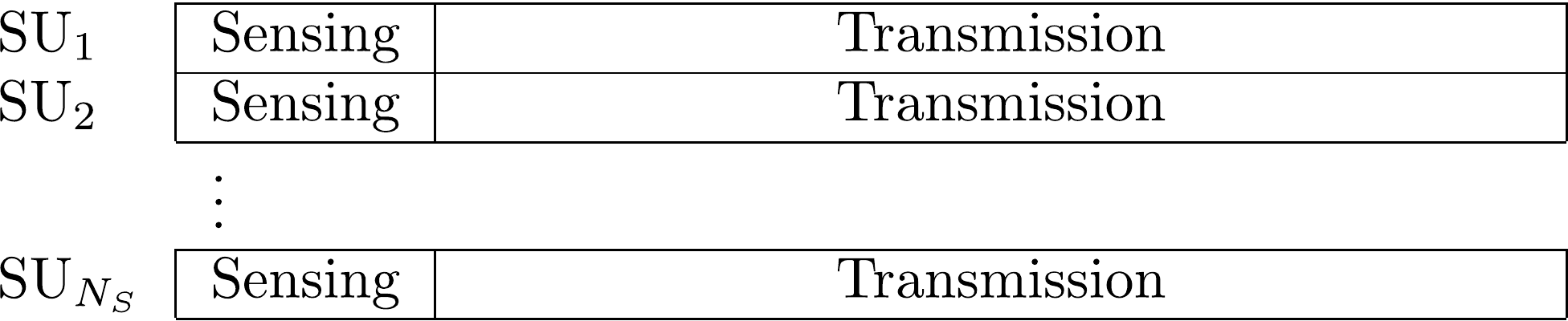}
        \caption{Time slotting of the SUs' operation. After sensing a particular subband the SUs send their local test statistics or decisions to the FC that makes the global decision about the state of the spectrum (sensing mini slot). Finally the FC grants permissions to transmit on the frequencies that were found to be idle (transmission slot).}
        \label{sensingSlots}
\end{figure}

The SUs are assumed to be synchronized and their operation to be divided into sensing mini time slots and potential transmission slots as illustrated in figure \ref{sensingSlots}. In a sensing time slot the SU senses up to $K_s$ subbands and then sends its local binary decision(s) to the FC via a dedicated control channel. The global decisions about the state of the sensed subbands is formed at the FC by combining the local binary decisions according a fusion rule. 

The FC may be a dedicated node or one or multiple nodes could serve as a FC in an ad hoc scenario. A dedicated FC makes a global decision on behalf of all other SUs, whereas individual FCs in an ad hoc scenario could make independent decisions based on their own test statistics and the test statistics received from other SUs. 

\begin{figure}[htp!]
        \centering      
        \includegraphics[width=0.5\columnwidth]{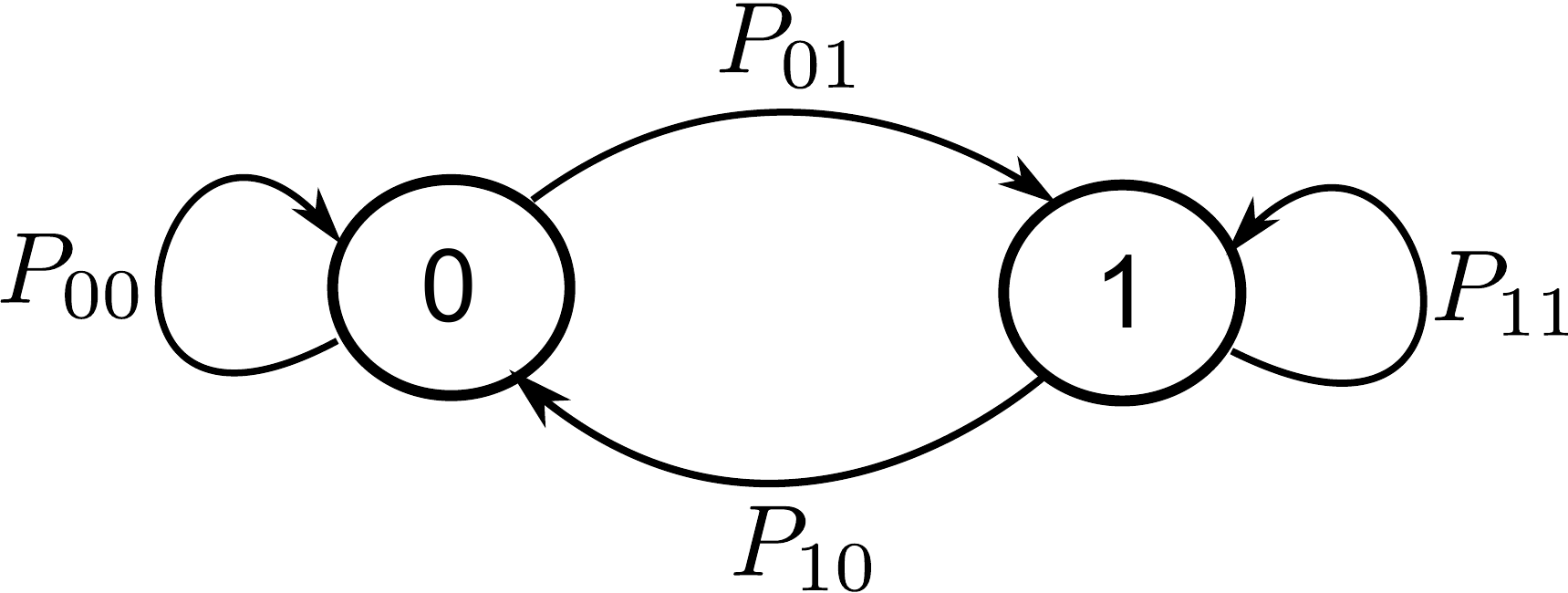} 
        \caption{The Gilbert-Elliot channel model \cite{gilbert}. In this paper state $0$ means that the subband is idle and state $1$ that the subband is occupied by a PU.}
        \label{markovModel}
\end{figure}

One proposed approach to model the PU activity is a two-state Markov chain shown in figure \ref{markovModel} \cite{zhao4}. In the model state $0$ means that the primary subband is idle (PU not transmitting) and state $1$ that the subband is occupied (PU transmitting). However, the policy proposed in this paper is not limited to the Markovian assumption. Markov model is merely used for illustration purposes in the experimental part of this paper.

\section{Reinforcement learning based sensing policy} \label{eGeedy}
In the PU network, as in most communication systems, the traffic load may vary depending on time and location. The expected amount of available radio spectrum for opportunistic secondary use may, for example, be much less during rush hours and in densely populated areas than during night time and in rural areas. Also the radio channel conditions fluctuate in time depending on location, velocity and frequency. Hence, the design of a sensing policy for CR has to be approached as a dynamic problem. 

\subsection{The $\epsilon$-greedy method}
Let $Q_{k}(a)$ denote the estimated value of action $a$ at time step $k$ and $a_{k}^{*}$ denote the selected action at time step $k$. The $\epsilon$-greedy policy is an ad-hoc method that balances between exploration and exploitation by selecting the action that has the highest estimated action value, i.e. $a_{k}^{*} = \arg\max_{a} Q_{k}(a)$, with probability $1-\epsilon$, or a random action, uniformly, with probability $\epsilon$ regardless of the action-value estimates~\cite{sutton98}.  

The $\epsilon$-greedy method is a simple and robust method that has minor computation and memory requirements. The random exploration phase allows replacing the random action selections with more carefully designed pseudorandom action selection with desired properties that are described in detail in section \ref{exploration}.
 
After taking action $a$ reward $r(a)$ is collected after which the Q-value of action $a$ is updated as \cite{sutton98}
\begin{equation}
\label{eq:value_update}
Q_{k+1}(a) = Q_{k}(a) + \alpha_k[r_{k+1}(a)-Q_{k}(a)],
\end{equation}
where $r_{k+1}(a)$ is the reward at time step $k+1$ for taking action $a$ and $\alpha_k$ $(0 < \alpha_k \leq 1)$ is a step size parameter. 

In a stationary scenario convergence is guaranteed with probability 1 when the step size parameter $\alpha_k$ satisfies the following conditions \cite{sutton98}
\begin{equation} \label{eq:conditions}
\sum_{k=1}^\infty \alpha_k = \infty \ \ \ \ \mathrm{and} \ \ \ \ \sum_{k=1}^\infty \alpha_k^2 < \infty.
\end{equation}
The first condition in \eqref{eq:conditions} guarantees that the step size is large enough to overcome the initial conditions, while the second condition guarantees that the step size is small enough to assure eventual convergence. Step size $\alpha_k = 1/(k+1)$ fulfills the conditions of \eqref{eq:conditions} and results in the standard sample-average of the past rewards. On the other hand, for constant $\alpha_k=\alpha$ the estimates will never completely converge, but continue varying in response to the latest observed rewards. In case of tracking a non-stationary process this is in fact desirable since the policy should react rapidly to the changes in subband occupancy statistics. A constant $\alpha_k=\alpha$ results in a weighted average of the observed rewards, i.e.~\cite{sutton98} 
\begin{equation} \label{eq:reward}
Q_{k+1}(a) = (1-\alpha)^{k+1}Q_{0}(a) + \sum_{i=1}^{k+1}\alpha(1-\alpha)^{k+1-i}r_{i}(a).
\end{equation} 
A constant step size $\alpha$ is suitable for tracking non-stationary processes such as the channel qualities in CR networks. It can be noticed in \eqref{eq:reward} that when $\alpha$ is large more emphasis is given on the most recent rewards whereas when  $\alpha$ is close to 0 the algorithm will give emphasis on rewards obtained in the more distant past as well. This suggests that for heavily non-stationary processes large values of $\alpha$ would be more suitable, whereas for stationary processes small $\alpha$ would give better results.

\subsection{The proposed sensing policy}
In this paper we propose a sensing policy using $\epsilon$-greedy exploration for selecting the frequency subbands to be sensed and for selecting the corresponding sensing assignments in a CR network. The policy is managed by the FC that tracks two kinds of Q-values: the Q-values for the subbands and the Q-values of all SUs to all subbands.
A natural way to define the reward $r_{k+1}(b)$ for selecting subband $b$ to be sensed is the obtained throughput:
\begin{equation}
r_{k+1}(b) = \begin{cases}
R_{k+1}(b), & \textrm{if $b$ is accessed and free} \\
0, & \textrm{if $b$ is occupied,}
\end{cases}
\end{equation}
where $R_{k+1}(b)$ is the instantaneous throughput on subband $b$. In this paper it is assumed that the SU who has been granted the permission to access the band will feed back an estimate of the achieved throughput. For example, this may be an estimate based on the measured channel quality between the communicating SUs. Using this feedback the FC updates the Q-values of each subband according to \eqref{eq:value_update}. 

The SU Q-values for particular subbands are updated by comparing the SUs' decision to the global decision: 
\begin{equation}
\label{eq:SU_reward}
r_{k+1}(s,b) = \begin{cases}
d_{k+1}(s,b), & d_{k+1}(\mathrm{FC},b) = 1 \\
Q_{k}(s,b), & d_{k+1}(\mathrm{FC},b) = 0,
\end{cases}
\end{equation}
where $d_{k+1}(s,b)$ denotes the local decision by SU $s$ for subband $b$ at time instant $k+1$ and $d_{k+1}(\mathrm{FC},b)$ denotes the corresponding decision at the FC. The SU's Q-value is then updated again according to \eqref{eq:value_update}. Hence, the SU's Q-value indicates its sensing performance at subband $b$, assuming that the global decision based on the local decisions from multiple SUs made at the FC is correct.

\begin{figure}[t!]
        \centering      \includegraphics[width=0.95\columnwidth]{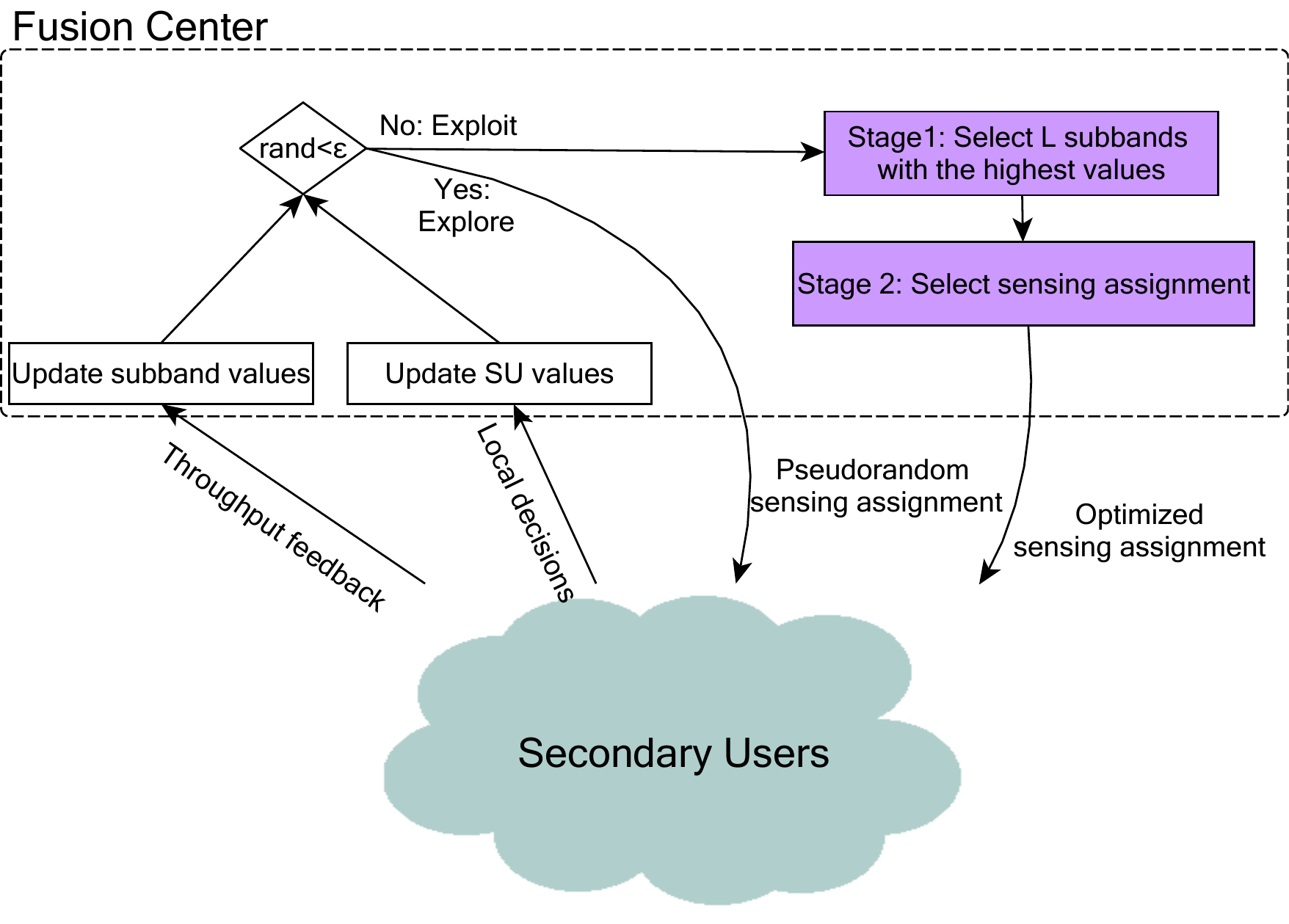}
        \caption{Flow diagram of the operation of the SU network and the FC. The blocks concerning the proposed sensing policy are highlighted with shading. In the diagram rand is uniformly distributed between $0$ and $1$.}
        \label{BlockDiagram_Bologna2}
\end{figure}

After all the Q-value updates, with probability $1-\epsilon$ the FC exploits its knowledge and selects $L$ subbands to be sensed that have the highest Q-values (stage 1 in figure \ref{BlockDiagram_Bologna2}). In this paper it is assumed that the FC has an estimate of the desired throughput and is able to select the parameter $L$ appropriately. After selecting the subbands the FC finds an appropriate sensing assignment for them (stage 2 in figure \ref{BlockDiagram_Bologna2}). With probability $\epsilon$ the sensing is done according to predefined pseudorandom frequency hopping codes with a fixed diversity order $D$, where $D$ is the number of SUs simultaneously sensing the same subband. In the exploitation phase the sensing assignment is the one that minimizes the number of sensings in the SU network while maintaining the detection performance at a desired level. Finally, the FC sends to the SUs information about which subbands they should sense.



\subsubsection{Exploration} \label{exploration}
\begin{figure}[htp!]
        \centering
\includegraphics[width=0.4\columnwidth]{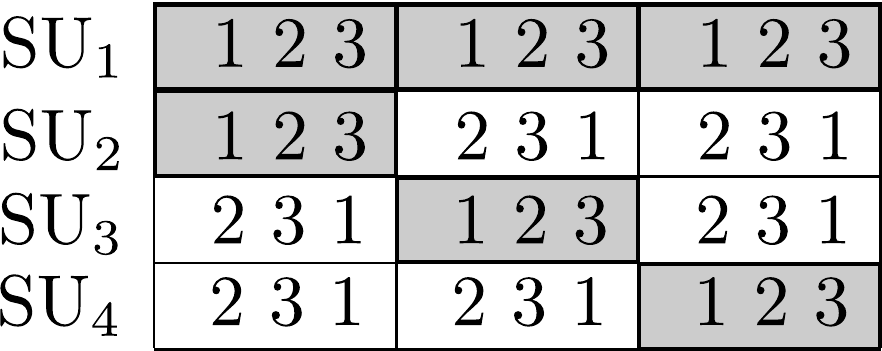}
        \caption{Pseudorandom frequency hopping codes for $N_S=4$, $N_B = 3$ and $D=2$. At each sensing instance the SU senses the subband pointed by the current entry of its hopping code. Hopping code entries are pointing to the physical frequencies that are maintained in a lookup table. The possible transmission slots following after each sensing instance have been dropped out for convenience.}
        \label{frequencyHoppingExample}
\end{figure}

In this section the pseudorandom frequency hopping based sensing policy proposed in \cite{oksanen1} is briefly summarized, since it constitutes the exploration phase of the sensing policy developed in this paper. The pseudorandom frequency hopping based sensing policy provides quick scanning of the spectrum of interest with minimal control signaling, thus being extremely suitable for exploring the spectrum. The frequency hopping code design allows for trading off scanning speed and diversity (and consequently detector performance) in an elegant manner. Moreover, by guaranteeing the desired diversity order $D$, reliable performance is ensured in demanding propagation environments.
In the pseudorandom frequency hopping based multi-band spectrum sensing policy the design of the sensing policy has been converted into designing and allocating pseudorandom frequency hopping codes to the SUs guiding them which subbands are sensed and when. After each hopping code period different $D$-tuples of the $N_S$ SUs will be employed to scan the spectrum of interest together. The design is made such that over time all possible SU combinations of size $D$ will be employed to sense each subband. Fig. \ref{frequencyHoppingExample} shows an example design of the hopping codes for $N_S=4$, $N_B=3$ and $D=2$. 

In frequency hopping based sensing each SU hops according to its hopping sequence to sense one of the subbands of interest. The subband to be sensed at time index $i$ is given by $f(i) = F[S_q(i)]$, where $S_q(i)$ is the $q$th frequency hopping sequence, $F$ is a table containing the mappings to the physical subbands. Table $F$ may include links to the subbands' center frequencies and bandwidths. It is assumed that $F$ is same for all SUs in the network. 

Since it is desirable to scan as much spectrum as possible at once, the hopping sequences are made orthogonal. The simplest way to generate an orthogonal code family is to cyclically shift any full sequence of integer numbers. A full sequence is a sequence that contains all integer numbers up to a certain number. Cyclic shifts may be generated by the modulo operation as
\begin{equation}
S_q(i) \equiv (i + \Delta_q) \mod N_B, 
\end{equation}
where $i \in [0,N_B-1]$, $q\in [0, \lfloor\frac{N_S}{D}\rfloor-1]$ and $\Delta_q$ is the shift parameter. For more information about the choice of $\Delta_q$ and the design of the frequency hopping sequences as well as simulation results see \cite{oksanen1}. 

\subsubsection{Exploitation} 
In many practical scenarios the cooperating SUs, although being in the vicinity of each others, may be in very different channel conditions due to fading.
Then the cooperation among the SUs may be optimized better in order to save energy of the SUs.

Assume that the secondary network of $N_S$ SUs wants to sense $L<N_B$ subbands in hope of spectral opportunities. These subbands have been selected in the first stage of the sensing policy as the ones that are most likely going to produce high reward (throughput) for the SU network. Denote the set of all the chosen L subband indices as $B$ and the set of all SU indices as $S$. Furthermore, assume that the SU network has knowledge about the SUs' probabilities of detection $P_{sb}$, where $s\in S$ and $b\in B$. In order to conserve the SUs' energy, we would like to minimize the number of SUs assigned for sensing while pursuing to guarantee a desired level of detection performance at the subbands of interest. Hence, the sensing assignment problem (SAP) can be formulated as
\begin{eqnarray} \label{SAP}
\min_{X} && \sum_{b\in B}\sum_{s\in S} w_{s}x_{sb}\\
\textnormal{s.t.} && \hat{P}_{miss,FC}^b(\mathbf{X}) \leq P_{miss,target}^b \nonumber \\
&& \sum_{b\in B} x_{sb} \leq K_s \nonumber \\
&& x_{sb} \in \{0,1\} \nonumber,
\end{eqnarray}
where $K_s$ is a positive integer corresponding to the number of subbands SU $s$ can sense simultaneously and $w_{s}$ is the weight of user $s$. $\mathbf{X}=[x_{sb}]$ is $N_S \times L$ the unknown binary sensing assignment matrix. The elements of $\mathbf{X}$ are
\begin{equation}
x_{sb} = 
\begin{cases}
        1 & , \textnormal{if SU } s \textnormal{ is assigned to sense subband } b\\
        0 & , \textnormal{otherwise}\\
\end{cases}.
\end{equation}
In equation \eqref{SAP} $\hat{P}_{miss,FC}^b(\mathbf{X})$ is the estimate of the miss detection probability at the FC at subband $b$ obtained with sensing assignment $\mathbf{X}$. $P_{miss,target}^b$ is the maximum probability of miss detection that the secondary network is allowed to have at band $b$. The first constraint in \eqref{SAP} requires that the probability of miss detection at the FC should be below the constraint, whereas the second constraint restricts the number of subbands SU $s$ can sense simultaneously to be $K_s$ or less. The weight $w_{s}$ of SU $s$ may be chosen, for example, according to the SUs' battery charge. If SU $s$ is known to have low battery charge it may be given relatively large weights compared to other users so that it will unlikely to be assigned for sensing. 

There are many ways to design distributed detection such that the detection performance constraint in \eqref{SAP} is met. As an example we consider here hard decision combining of multiple Neyman-Pearson detectors \cite{Varshney1996}. Neyman-Pearson detectors maximize the detection probability under a constraint on false alarm rate and hence false alarm rate is not included in \eqref{SAP} as a separate constraint. Typically the false alarm rate constraint is set small since false alarms equal to overlooked spectral opportunities. For hard decision combining, such as the OR-rule considered in the next subsection, the false alarm rate constraint at the FC is simply met by controlling the local detection thresholds according to the number of SUs assigned to sense the same subband \cite{Varshney1996}. 

\subsubsection{Sensing assignment for the OR-fusion rule}
Next the sensing assignment is illustrated for the OR-rule where the SUs send only their local decisions to the FC. The FC then decides a subband to be free only if all sensing SUs have reported it to be free. Other fusion rules such as $K$-out-of-$N$-rule could be used as well. Assuming conditional independence of the observations at different SUs given $H_0$ or $H_1$ the probability of missed detection at the FC at subband $b$ for the OR-rule is given by 
\begin{equation}
P_{miss,FC}^b=\prod_{s=1}^{N_s} (1-P_{sb})^{x_{sb}},
\end{equation}
which as such would lead to a nonlinear constraint in the SAP given by equation \eqref{SAP}. However, the detection performance constraint can be linearized by simply taking the logarithm of the missed detection probabilities, i.e. 
\begin{equation}
\ln(P_{miss,FC}^b)=\sum_{s=1}^{N_s} \ln(1-P_{sb})x_{sb}.
\end{equation}
Then the SAP for the OR-rule can be formulated as a linear binary integer programming (BIP) problem as
\begin{eqnarray} \label{SAP_OR}
\min_{x} && \mathbf{w}^T\mathbf{x} \\
\textnormal{s.t.} && \mathbf{A}\mathbf{x}\leq \mathbf{c}\nonumber \\ 
&& \textnormal{$\mathbf{x}$ is binary},\nonumber
\end{eqnarray}  
where $\mathbf{w}$ is an $N_SL \times 1$ vector of weights for the SUs at different subbands, $\mathbf{x}=\textnormal{vec}(\mathbf{X})$ is a binary vector of size $N_SL\times 1$, $\mathbf{A}$ is the $(L+N_S)\times LN_S$ constraint matrix containing the logarithms of the estimated local miss detection probabilities $\ln(1-\hat{P}_{sb})$'s and $L$ identity matrices $\mathbf{I}_{N_S}$ at the bottom and $\mathbf{c}$ is the vector of the constraints. The constraint vector is given as $\mathbf{c} = [\ln(P_{miss,target}^1),...,\ln(P_{miss,target}^{L}),K_1,...,K_{N_S}]^T$. Since the detection probability can be known only up to a certain margin of error, the constraint vector $\mathbf{c}$ should in practice include a safety margin defined by a spectrum regulator. The constraint matrix $\mathbf{A}$ is given by
\begin{equation}
\mathbf{A}=\left[\begin{tabular}{ccccccccc}
$\hat{\mathbf{p}}_{miss}^1$ & 0  & $\cdots$  \\ 
 & $\hat{\mathbf{p}}_{miss}^2$ & 0 & $\cdots$ \\ 
 &  & $\ddots$ &  \\ 
 & $\cdots$ & 0 & $\hat{\mathbf{p}}_{miss}^{L}$ \\ 
$\mathbf{I}_{N_S}$ & $\mathbf{I}_{N_S}$ & $\cdots$ & $\mathbf{I}_{N_S}$ \\ 
\end{tabular}\right],
\end{equation}
where $\hat{\mathbf{p}}_{miss}^b = [\ln(1-\hat{P}_{1b}), \ln(1-\hat{P}_{2b}),..., \ln(1-\hat{P}_{N_sb})]$ and $\mathbf{I}_{N_S}$ is the identity matrix of size $N_S$.

This BIP problem is NP-hard but solvable by branch-and-bound (BB) type algorithms. The worst case running time of BB search, although unlikely, is $2^{N_SL}$, that corresponds to the case where no branching is possible. In practice $N_SL$ maybe assumed to be small. 

In cases where the product $N_SL$ is large, the probability that there exists multiple near optimal assignments is high. In such cases heuristic approximation algorithms may be applied. In \cite{wang2011} an iterative Hungarian algorithm is proposed to find a sensing assignment that minimizes the probability of miss detection. The policy assigns SUs to sense the subbands one by one using the Hungarian method \cite{kuhn}. In our problem formulation, the Hungarian method can be employed iteratively, similarly to \cite{wang2011}, to find a near optimal solution for the SAP with $w_{s} = 1$ by modifying the algorithm to stop immediately once a feasible solution is found.  Since the Hungarian algorithm runs in polynomial time, this method is also polynomial time.

\subsection{SU Q-value and the local detection probability}
Solving the optimization problem of \eqref{SAP} requires the estimates of the probabilities of missed detection at the FC.
Defining the reward as in \eqref{eq:SU_reward} provides simultaneously a simple estimate for the SUs' probabilities of detection. 

Since the SU Q-values are updated according to equation \eqref{eq:value_update} similarly to the subband Q-values, it can be shown that the asymptotic expected Q-values $\mathrm{E}[Q_k(s,b)]$ approach the expected reward as $k\rightarrow \infty$. From equation \eqref{eq:SU_reward} assuming that $\mathrm{E}[Q_{k+1}(s,b)] \displaystyle{\mathop{=}_{k\rightarrow\infty}} \mathrm{E}[Q_{k}(s,b)]$ we get
\begin{eqnarray*}
\lim_{k\rightarrow\infty}\mathrm{E}[Q_{k+1}(s,b)] =  \lim_{k\rightarrow\infty}\mathrm{E}[r_{k}(s,b)] &=& \\
\frac{P_1\mathrm{P}(d(FC)=1 \cap d(s)=1|H_1)}{P_1P_{d,FC} + P_0P_{f,FC}} &+&\\ \frac{P_0\mathrm{P}(d(FC)=1 \cap d(s)=1|H_0)}{P_1P_{d,FC} + P_0P_{f,FC}}&,&
\end{eqnarray*}
where $P_0$ is the probability of the subband being free, $P_1=1-P_0$, $d(s)$ and $d(FC)$ are the decision at SU $s$ and at the FC, respectively, and $P_{d,FC}$ and $P_{f,FC}$ are, respectively, the probabilities of detection and false alarm at the FC. For notational convenience the subband index $b$ has been dropped. 

For the OR-rule 
$\mathrm{P}(d(FC)=1 \cap d(s)=1|H_1) = \\ \mathrm{P}(d(s)=1 | H_1) = P_{d,s}$ and $\mathrm{P}(d(FC)=1 \cap d(s)=1|H_0) = \mathrm{P}(d(s)=1 | H_0) = P_{f,s}$, since $\mathrm{P}(d(FC)=1 | d(s)=1)=1$. Then,
\begin{equation*}
\lim_{k\rightarrow\infty}\mathrm{E}[Q_{k+1}(s,b)] = \frac{P_1P_{d,s} + P_0P_{f,s}}{P_1P_{d,FC} + P_0P_{f,FC}}\approx \frac{P_{d,s}}{P_{d,FC}},
\end{equation*}
assuming that $P_0P_{f,s}\approx 0$ and $P_0P_{f,FC}\approx 0$.
It can be seen that in order for the local detection probability estimates to be close enough to the detection probability at the FC $P_{d,FC}$ should be close to one. This can be achieved through spatial diversity if the decision at the FC is based on multiple SUs' local test statistics or decisions.

\begin{figure}[t!]
        \centering      \includegraphics[width=0.7\columnwidth]{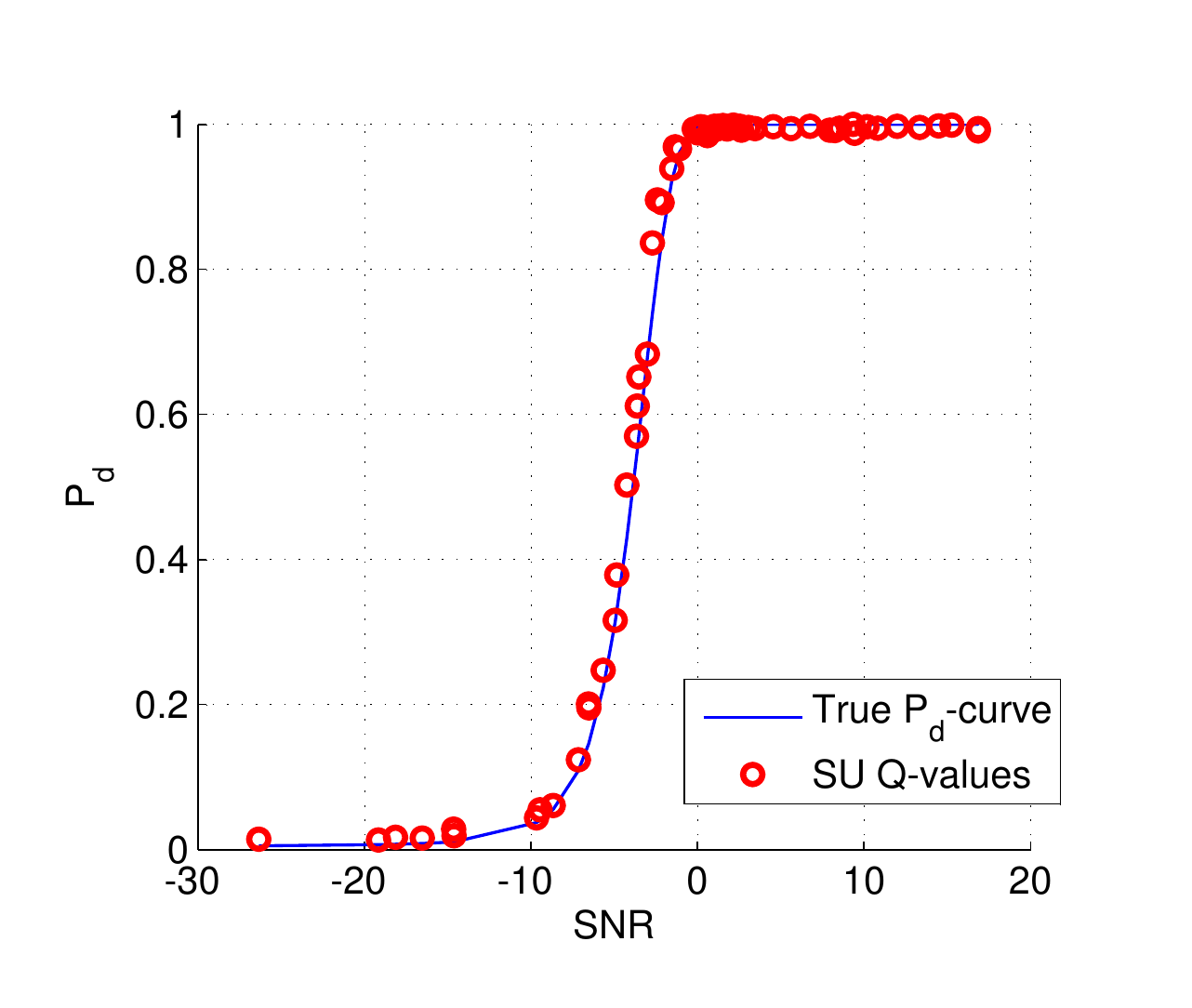}
        \caption{The mean converged SU Q-values after 10000 sensing instances ordered according to the corresponding mean SNR and the true probability of detection curve. The mean was calculated over 100 runs. It can be seen that the SUs' Q-values (red circles) coincide well with the true probabilities of detection (blue curve).}
        \label{SU_Qvalues_NCjournal}
\end{figure}

Figure \ref{SU_Qvalues_NCjournal} shows converged SU $Q$-values ordered by mean SNR and the true probability of detection curve. The detection scheme is Neyman-Pearson energy detection with a sample size $50$ and $P_{f,FC}=0.01$. The fusion rule is the OR-rule with $D=2$ and $\alpha=0.1$. It can be seen that the Q-values align with the true probabilities of detection. 

\subsection{Convergence of the subband Q-values} \label{QvalueConvergence}
Since all the subbands are not necessarily sensed all the time, we need to introduce another time variable $T_k(b)\leq k$ denoting the number of sensing instances (and value updates) at band $b$ up to the $k$th run of the $\epsilon$-greedy algorithm. Regrouping the components in equation \eqref{eq:value_update} the Q-value of subband $b$ can be expressed as
\begin{equation}
Q_{T_k(b)+1}(b) = (1-\alpha)Q_{T_k(b)}(b) + \alpha r_k(b). \nonumber
\end{equation}
Taking the expectation of both sides results to
\begin{equation}
\textrm{E}[Q_{T_k(b)+1}(b)] = (1-\alpha)\textrm{E}[Q_{T_k(b)}(b)] + \alpha\mu(b), \nonumber
\end{equation}
where $\mu(b)=\textrm{E}[r_k(b)]$. This is a linear recurrence, whose solution is given by 
\begin{equation}
\textrm{E}[Q_{T_k(b)+1}(b)] = \alpha^{T_k(b)+1}\textrm{E}[Q_{0}(b)] + \nonumber \left(1-(1-\alpha)^{T_k(b)+1}\right)\mu(b).
\end{equation}
Assuming $\textrm{E}[Q_{0}(b)]=0$ the expected Q-value of band $b$ at the $T_k(b)$th update is
\begin{equation}
\textrm{E}[Q_{T_k(b)}(b)] = (1-(1-\alpha)^{T_k(b)})\mu(b) = \mu(b). \nonumber
\end{equation}
as $T_k(b)\rightarrow\infty$.
Then the expected Q-value of band $b$ after the $k$th run of the $\epsilon$-greedy algorithm is given by
\begin{equation} \label{theorConvergence}
\textrm{E}[Q_{k}(b)] = \mu(b)\sum_{T_k(b)=0}^{k}\textrm{P}(T_k(b))(1-(1-\alpha)^{T_k(b)}),
\end{equation}
where $\textrm{P}(T_k(b))$ is the probability that band $b$ has been updated $T_k(b)$ times within the $k$ runs of $\epsilon$-greedy algorithm.

Upper and lower bounds can be easily obtained for $\textrm{P}(T_k(b))$ in a stationary case: 
\begin{equation}
\mathcal{B}(T_k(b),k,\epsilon\frac{L}{N_B}) \leq \textrm{P}(T_k(b)) \leq \mathcal{B}(T_k(b),k,1-\epsilon(1-\frac{L}{N_B})) \nonumber,
\end{equation}
where $\mathcal{B}(T_k(b),k,p) = \binom{k}{T_k(b)}p^{T_k(b)}(1-p)^{k-T_k(b)}$ is the binomial probability density function. The lower bound corresponds to the probability that the Q-value is updated only in the exploration phase and the upper bound to the case that in the exploitation phase the Q-value is updated with probability one. 

The analysis for the convergence of the SU Q-values is almost identical to the analysis above for the Q-values of the subbands. The probability $\textrm{P}(T_k(b,s))$ that SU $s$ has sensed subband $b$ during $k$ runs $T_k(b,s)$ times is then bounded as
\begin{equation}
\mathcal{B}(T_k(b,s),k,\epsilon\frac{L}{N_B^2}) \leq \textrm{P}(T_k(b,s))\nonumber \leq \mathcal{B}(T_k(b,s),k,1-\epsilon(1-\frac{L}{N_B^2})).
\end{equation}
Establishing a lower bound for the probability of the number of sensings is important for guaranteeing a desired convergence rate for the estimates of the probability of missed detections in the second stage of the proposed sensing policy.

\begin{figure}[t!]
        \centering      \includegraphics[width=0.75\columnwidth]{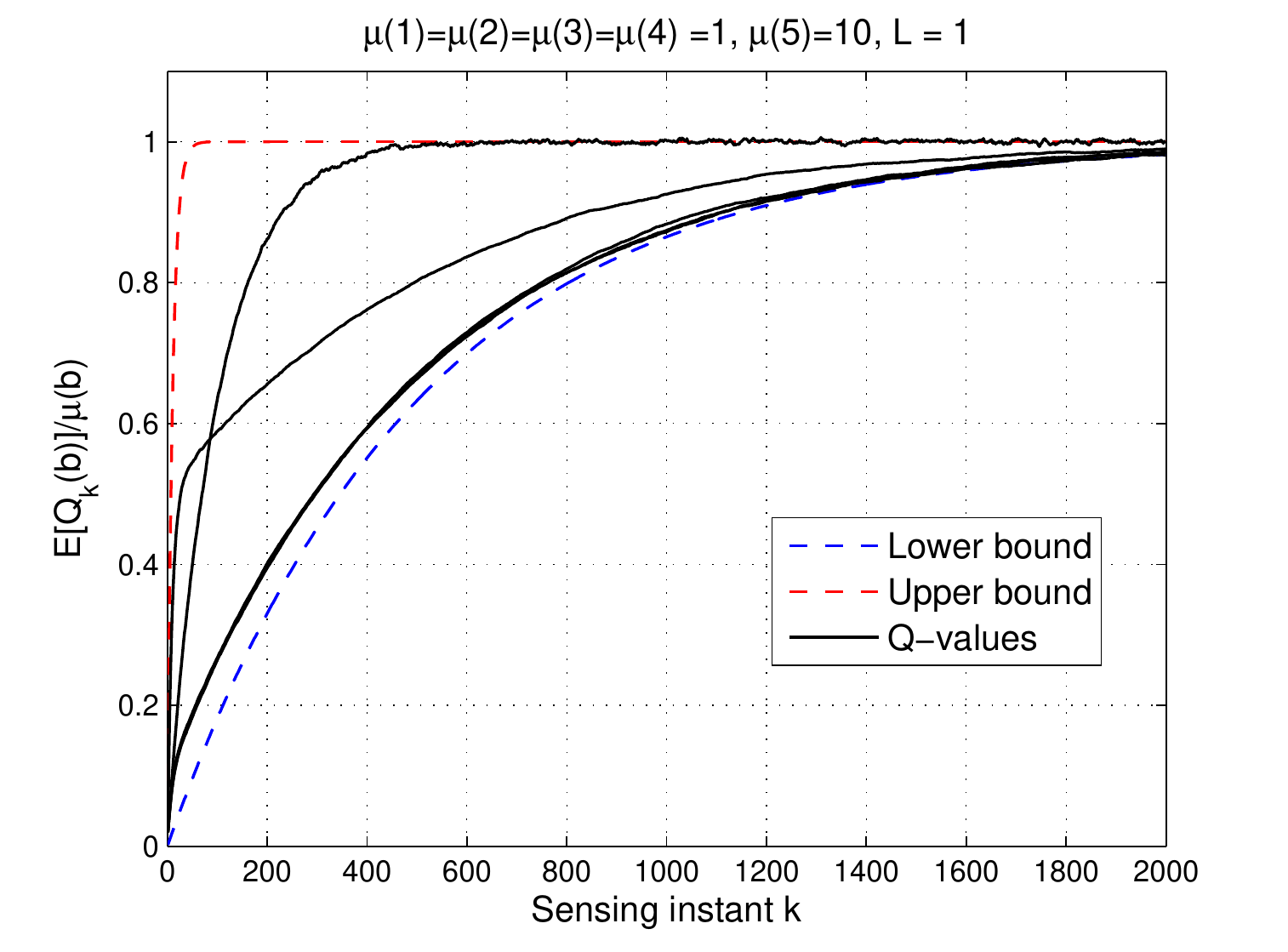}
        \caption{The simulated convergence of the expected Q-value (black solid curves) of 5 subbands and the upper (red dashed curve) and lower bounds (blue dashed curve) for them. The number of sensed bands is set to $L=1$. The rewards are assumed to be Bernoulli distributed with probability 0.5 and means $\mu(1)=\mu(2)=\mu(3)=\mu(4) =1$ and $\mu(5)=10$. The used parameters are $\epsilon = 0.1$ and the step size $\alpha=0.1$. It can be noticed that the convergence of the Q-values of the subbands with the lowest mean rewards follows closely the lower bound, where as the convergence at the best subband goes closer to the upper bound.}
        \label{expectedQ}
\end{figure}

Figure \ref{expectedQ} shows the simulated convergence of the expected Q-values of 5 subbands and the upper and lower bounds for them. The number of sensed bands is set to $L=1$. The rewards are assumed to be Bernoulli distributed with probability $P_0=0.5$ and means $\mu(1)=\mu(2)=\mu(3)=\mu(4) =1$ and $\mu(5)=10$.

\section{Simulation examples} \label{simulations}
In this section simulation results for the proposed sensing policy are shown. The main focus is put on the obtained throughput of the secondary network and miss detection probability.

\subsection{Stationary case}
This subsection provides the results for a stationary scenario in which the occupancy statistics of the primary bands stay constant during the whole simulation period. The results are shown for the throughput, average miss detection probability and relative number of sensings in the SU network with different values of $\epsilon$. Furthermore, the simulations are shown for comparison using the exact BB search and an approximative iterative Hungarian (IH) method adapted from \cite{wang2011}. In the stationary case the mean detection performances of the SUs remain constant. The simulations are done for $N_S=6$ SUs and $N_P=10$ primary subbands. The availability of each subband is modeled according to a two state Markov chain (see figure \ref{markovModel}) with state transition probabilities $P_{11}=P_{00}=0.9$. Different subbands are assumed to be independent of each other. The mean SNRs of the primary signal in the secondary network is assumed to be distributed according to the log-normal shadow model with a standard deviation of 9 dBs. The fast fading component of the channel is modeled as a block fading Rayleigh channel with expected power gain of $1$. Furthermore, it is assumed that $3$ of the subbands are able to provide $10$ times higher throughputs on average. For spectrum sensing Neyman-Pearson energy detection with a sample size of $50$ is used. The global decisions at the FC are formed using the hard decision OR-rule with a constant false alarm rate $P_{f,FC}=0.01$. In the exploration phase the pseudorandom frequency hopping code design is made using a fixed diversity order $D=2$ that has been selected such that on average the desired miss detection probability is close to $P_{miss,target}$. In the exploitation phase the number of subbands SUs can sense simultaneously is set to $K_{s} = 1, \forall s \in S$,  and the target probability of miss detection at the subbands $P_{miss,target}(b)=0.1$. The weights $w_s$ in the SAP have been set to 1 for all SUs. The number of subbands that the SU network wants to find is constant during the whole simulation, i.e., $L=3$. For clarity in this section the step sizes in the first and seconds stage of the sensing policy are denoted as $\alpha_1$ and $\alpha_2$ respectively. In the simulations $\alpha_1 = 0.01$ and $\alpha_2=0.1$.

\begin{figure}[t!]
        \centering 
            \includegraphics[width=0.75\columnwidth]{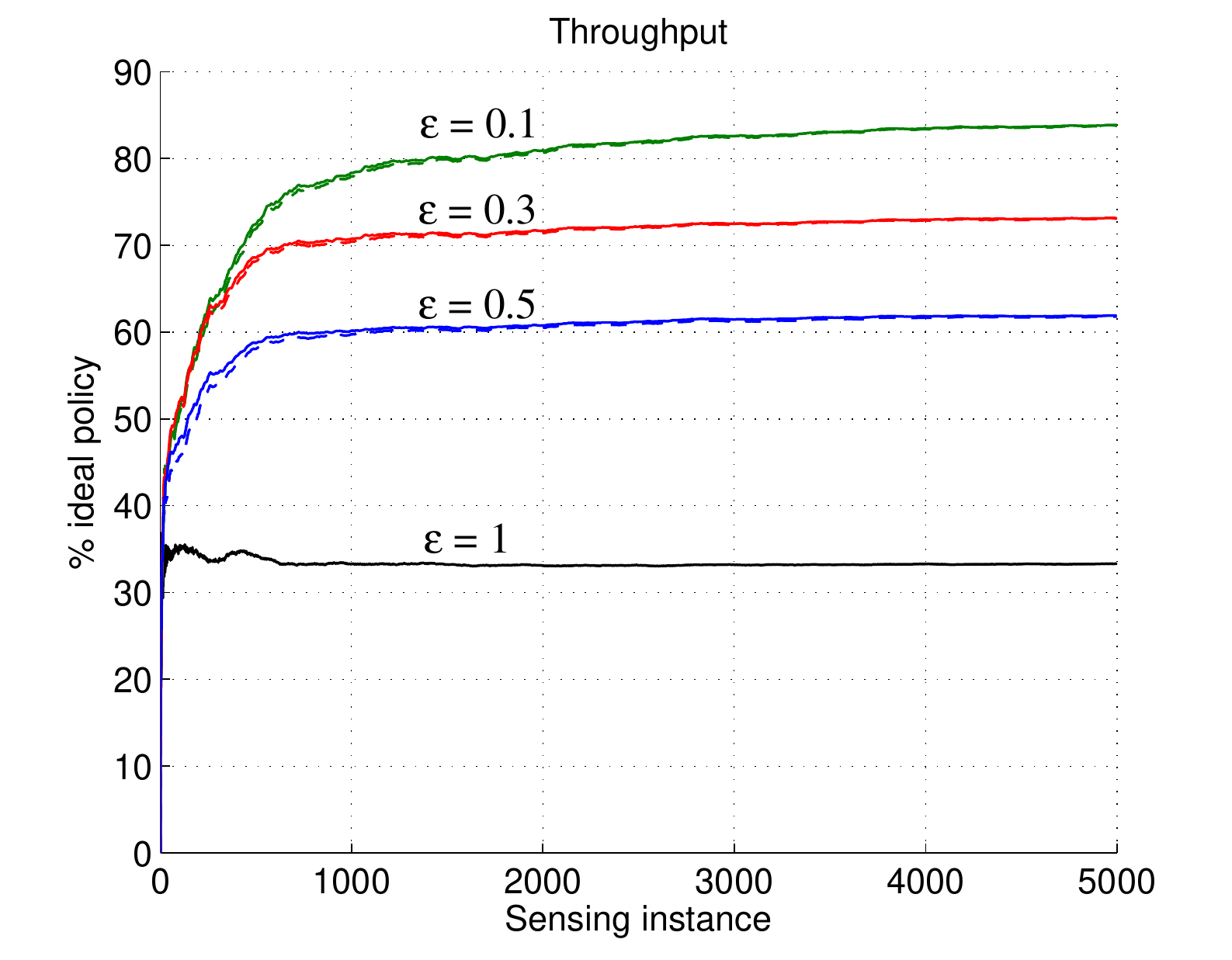} 
        \caption{Cumulative throughput relative to an ideal policy. The dashed curves are with the IH method and the solid curves with the exact BB algorithm. For $\epsilon=1$ the curves are exactly the same as it corresponds to exploration only case. For $\epsilon<1$ cases the performances are almost the same. For example with $\epsilon=0.1$ the proposed policy is able to provide about 83\% of the throughput of an ideal policy. It can be seen that with small $\epsilon$ the converged throughput is high, whereas the convergence rate is slow. 
        }
        \label{TPall_NCjournal}
\end{figure}

Figure \ref{TPall_NCjournal} shows the cumulative throughput relative to an ideal, genie aided policy. An ideal policy is assumed to be able to find all spectrum opportunities and select the $L$ subbands with highest instantaneous throughputs. The obtained throughput using the exact BB search and the throughput using the heuristic IH method are practically the same. However, in this case the IH method found the assignment on average 80 times faster than the BB search. It can be noticed that with $\epsilon=0.1$ the proposed policy is finally obtaining 83\% of the throughput of an ideal policy. As can be seen the trade-off with small $\epsilon$ comes naturally with a slower rate of convergence.

\begin{figure}[t!]
        \centering     
         \includegraphics[width=0.75\columnwidth]{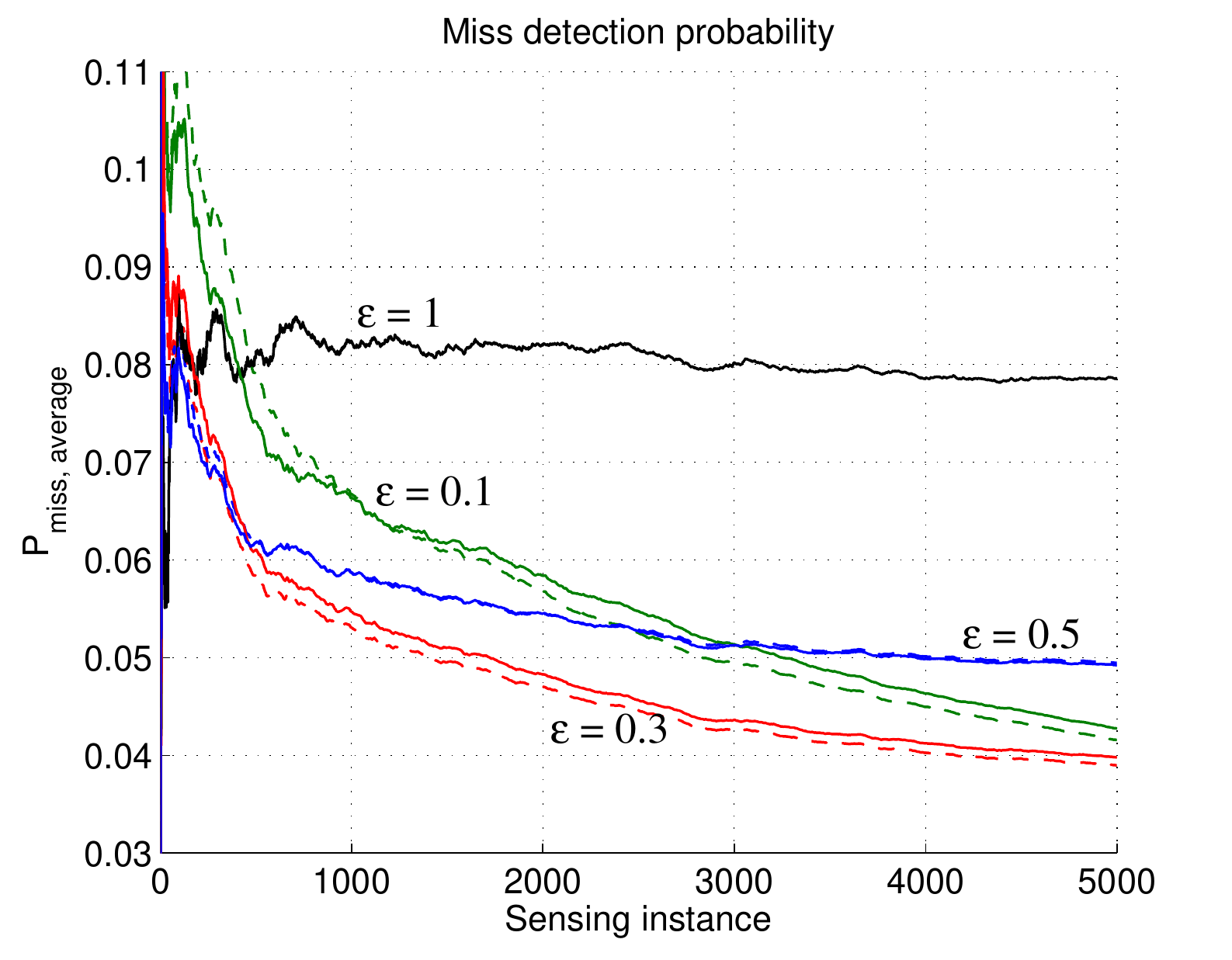}
        \caption{Convergence of the missed detection probability for different choices of $\epsilon$ averaged over the subbands. The dashed curves are with the IH method and the solid curves with the exact BB algorithm. For $\epsilon=1$ the curves are exactly the same as it corresponds to exploration only case. For $\epsilon<1$ cases the performances are almost the same. It can be seen that over time for $\epsilon< 1$ the missed detection probability converges below $P_{miss,target}=0.1$. Again for small $\epsilon$ the converging value is compromised with slow convergence rate. 
        }
        \label{PMall_NCjournal}
\end{figure}

Figure \ref{PMall_NCjournal} shows the probability of miss detection for different choices of $\epsilon$. The diversity order for the fixed policy (curve corresponding to $\epsilon=1$) was selected such that on the average the target miss detection probability is achieved. The resulting average miss detection probability using the exact BB search and using the heuristic IH method are almost the same. When $\epsilon$ is decreased the policy starts assigning those SUs with high probability of detection to sense the corresponding subbands more often thus decreasing the overall number of miss detections. For $\epsilon=0.1$ and $\epsilon=0.3$ the average miss detection probability is finally at the end of the simulation close to 0.04. 

\begin{figure}[t!]
        \centering     
         \includegraphics[width=0.75\columnwidth]{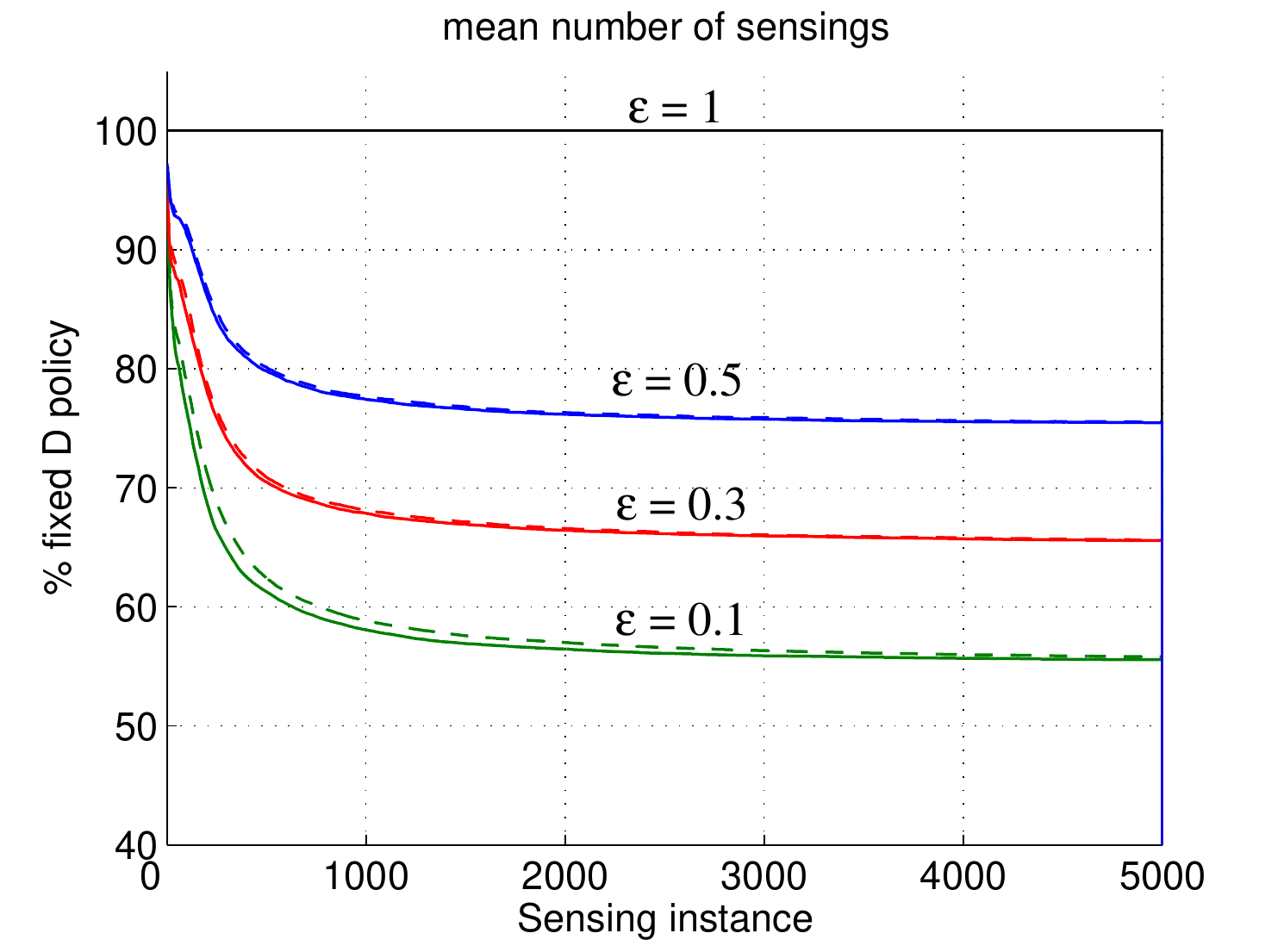}
        \caption{Number of sensings relative to a sensing policy with fixed diversity order $D=2$. The dashed curves are with the IH method and the solid curves with the exact BB algorithm. For $\epsilon=1$ the curves are exactly the same as it corresponds to exploration only case. For $\epsilon<1$ cases the performances are almost the same. It can be seen that the proposed policy reduces the number of sensings in the secondary network by assigning only the best SUs for sensing. With $\epsilon=0.1$ the number of sensing SUs per subband is almost halved, meaning that on average in the exploitation phase there is only one SU sensing per subband. 
        }
        \label{Eall_NCjournal}
\end{figure}

Figure \ref{Eall_NCjournal} shows the number of sensings over time compared to a sensing policy with fixed diversity order $D=2$ (exploration only). The savings in the number of sensings using the exact BB search and using the heuristic IH method are again practically same. For the case $\epsilon=0.1$ the number of sensings and transmissions of the local sensing results to the FC are reduced to 56\%. 

\subsection{Expected throughput for non-stationary cases}
For a non-stationary scenario the throughput of the proposed sensing policy is compared against two other methods. The results are shown only for the first stage of the proposed sensing policy that attempts to maximize the throughput of the secondary network. The results are shown for a case in which the availability of the subbands is Markov process and for a case in which the availability is a Bernoulli process (i.e. a special case of a two-state Markov chain). Moreover, the proposed policy is compared to two other state-of-the-art policies. Namely, the comparison is done against the discounted UCB (DUCB) policy with a discount factor $\gamma$ \cite{kocsis} and a near-optimal sensing policy \cite{liu3}, the Whittle index policy, that assumes the state transition probabilities in the Markov chain to be known. The comparison between the Whittle index policy and the two machine learning-based policies is therefore not entirely fair since the assumptions about prior knowledge are different.

Here the number of subbands has been set to $N_B=5$ and the number of simultaneously sensed bands to $L=1$. The missed detection probability at the FC is assumed to be $P_{miss,FC}=0.1$ and the false alarm rate $P_{f,FC}=0.01$ using Neyman-Pearson detectors. The mean throughputs of the bands bands are $[11, 21, 31, 41, 51]$. In the first scenario the transition probabilities of the Markov chain are initialized as $P_{00} = [0.5, 0.9, 0.6, 0.8, 0.8]$ and $P_{11} = [0.9,0.31,0.7,0.9,0.3]$. In the Bernoulli case the probabilities of the subbands being free are initialized as $P_0 = [0.87, 0.17, 0.43, 0.33, 0.78]$. To simulate non-stationary behavior the transition probabilities and the mean rewards are randomly permutated among the subbands at random time instances.

\begin{figure}[t!]
        \centering      \includegraphics[width=0.8\columnwidth]{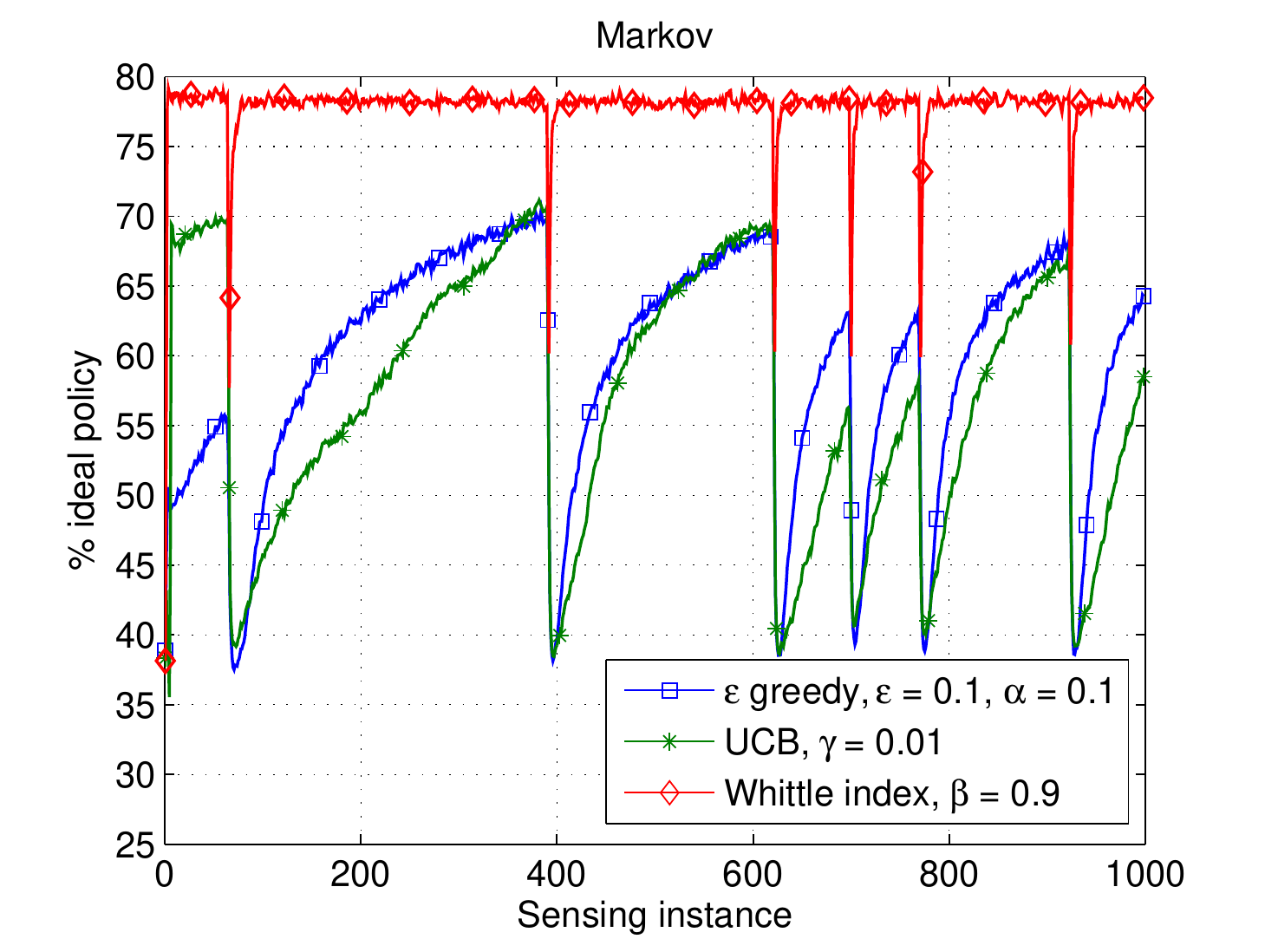}
        \caption{The sum throughput of the secondary network over time in a non-stationary scenario with Markovian rewards. The throughput statistics are permutated randomly among the subbands at randomly selected time instants that show up as sudden deep drops in the expected throughput. It can be seen that the DUCB provides high throughput in the beginning when all the discounted mean throughputs in the algorithm are zero. However, the proposed sensing policy with $\epsilon$-greedy exploration seems to provide more stable convergence at all times. Naturally the Whittle index based policy has the best convergence at all times, since it is assumes that the throughput distributions are known.}
        \label{TpAll_markov_1}
\end{figure}

Figure \ref{TpAll_markov_1} shows the expected mean throughput for the non-stationary Markovian case. Since it is assumed that the Whittle index policy knows the throughput distributions perfectly at each time, the optimized policy is naturally giving the highest throughput. It can be seen that DUCB adapts fast at the beginning when the discounted mean throughputs in the algorithm have been set to zero. However, after the first change in the throughput distributions the convergence of DUCB slows down significantly. The proposed sensing policy with $\epsilon$-greedy exploration seems to provide more consistent convergence at all times. 

\begin{figure}[htp!]
        \centering      \includegraphics[width=0.8\columnwidth]{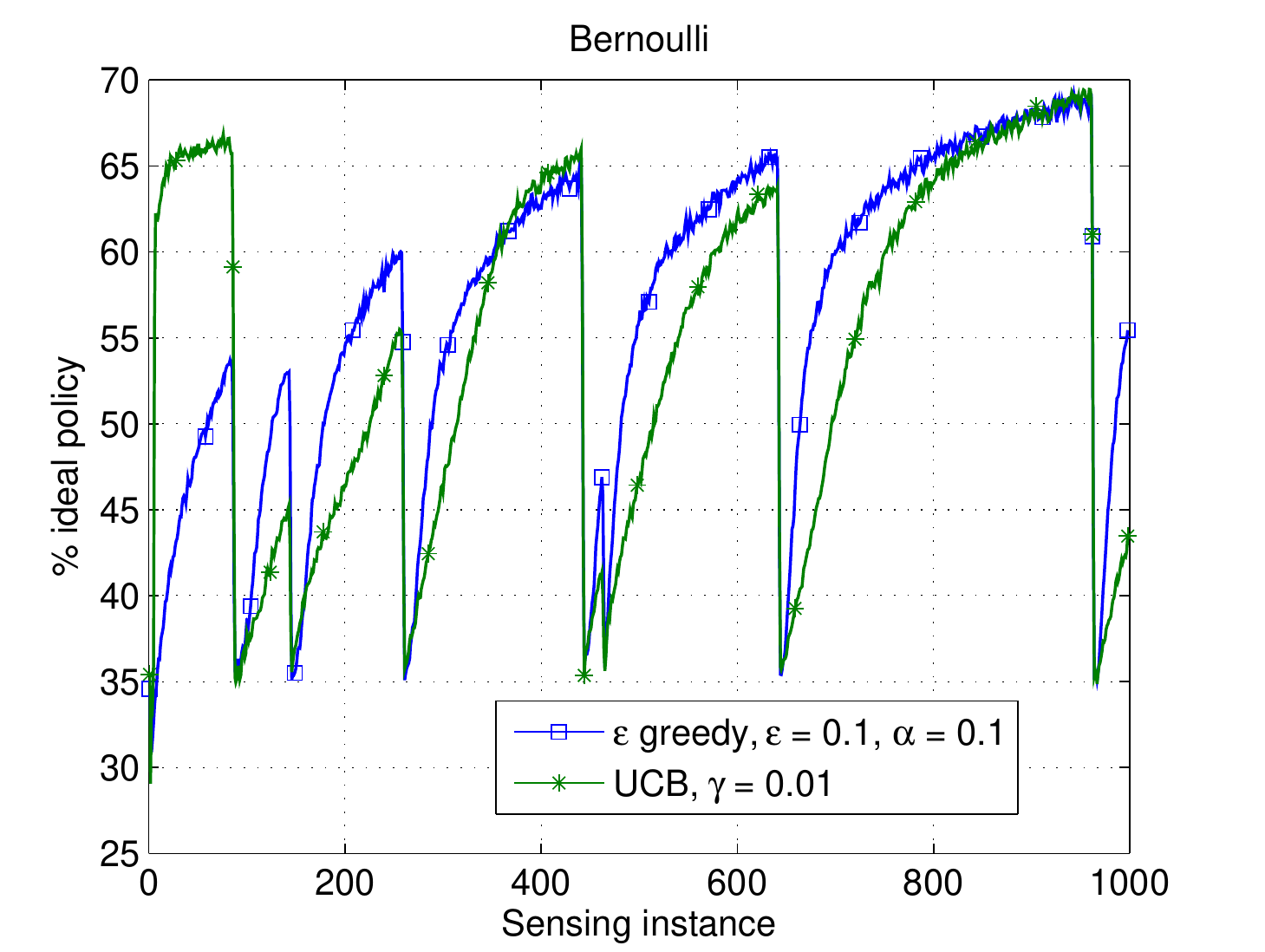}
        \caption{The sum throughput of the secondary network over time in a non-stationary scenario with Bernoulli rewards. The throughput statistics are rotated among the subbands randomly at randomly selected instances that show up as sudden deep drops in the expected throughput. The DUCB performs again well in the beginning but degrades notably after the first   change in the throughput distributions.}
        \label{TpAll_Bernoulli_1}
\end{figure}

Figure \ref{TpAll_Bernoulli_1} shows the expected throughput for the non-stationary Bernoulli case. Here only results are shown for the two machine learning-based policies, since neither of them does not assume any prior knowledge about the underlying Bernoulli process. The two machine learning based sensing policies perform almost alike as in the first non-stationary scenario, with the proposed policy giving slightly better overall performance than the DUCB policy.  

\section{Conclusions} \label{conclusions}
In this paper a machine learning based multi-band spectrum sensing policy is proposed. In the proposed policy the $\epsilon$-greedy method is employed to track the occupancy statistics of the PU and to estimate the detection performance of the SUs. 
Using the $\epsilon$-greedy method the proposed policy exploits the gained knowledge about the throughputs of different subbands by selecting the sensed subbands as the ones with the highest Q-value. Furthermore, knowledge about the detection performances of different SUs is exploited by minimizing the number of SUs assigned for sensing that are collaboratively able to meet a desired miss detection probability threshold. 

Exploration of the radio spectrum and different sensing assignments is realized using pseudorandom frequency hopping codes with fixed diversity order. Firstly, the pseudorandom exploration with fixed diversity order guarantees reliable sensing, and secondly, eventually all possible SU combinations of size $D$ will be considered. 

In the exploitation phase the sensing assignment problem is formulated as a binary integer programming problem in which the objective is to minimize the number of sensors $D$ per subband while ensuring the desired detection performance at each subband. By minimizing the number of sensing SUs per subband energy of the battery operated users is conserved and the amount of transmitted local test statistics is reduced. The optimal sensing assignment may be found by using exact branch-and-bound search or an approximative algorithm such as the iterative Hungarian method.

In this paper we demonstrate the performance of the proposed sensing policy and derive analytical expressions about the convergence of the policy. The simulation results show that the proposed sensing policy provides excellent performance in terms of throughput, detection probability and energy efficiency. 

%
\bibliographystyle{elsarticle-num-names}
\bibliography{viitteet}

\end{document}